\documentclass[11pt,a4paper]{article}
\usepackage[hyperref]{acl2019}
\usepackage[T1]{fontenc}
\usepackage[utf8]{inputenc}

\usepackage{times}
\usepackage{latexsym}
\usepackage{graphicx} 
\usepackage{array} 
\usepackage{booktabs} 
\usepackage{url}
\usepackage{multirow}

\usepackage{scalerel,stackengine}
\stackMath
\newcommand\reallywidehat[1]{%
\savestack{\tmpbox}{\stretchto{%
  \scaleto{%
    \scalerel*[\widthof{\ensuremath{#1}}]{\kern.1pt\mathchar"0362\kern.1pt}%
    {\rule{0ex}{\textheight}}
  }{\textheight}%
}{2.4ex}}%
\stackon[-6.9pt]{#1}{\tmpbox}%
}

\usepackage[normalem]{ulem}

\usepackage{siunitx} 
\aclfinalcopy 
\def\aclpaperid{***} 

\usepackage{makecell}

\newcommand{\insertEvaluationDatasetTable}{
    \begin{table}[t]
        \begin{center}
            \small
            \begin{tabular}[b]{cc|l}
            \toprule
             & & \textbf{Evaluation Dataset} \\
            \midrule
            \multicolumn{2}{l|}{\textsc{en-cs}}       & newstest2018 \\
            \multicolumn{2}{l|}{\textsc{en-de}}       & newstest2018 \\
            \multicolumn{2}{l|}{\textsc{en-fi}}       & newstest2018  \\
            \multicolumn{2}{l|}{\textsc{en-gu}}       & newsdev2019 \\
            \multicolumn{2}{l|}{\textsc{en-kk}}       & newsdev2019 \\
            \multicolumn{2}{l|}{\textsc{en-lt}}       & newsdev2019 \\
            \multicolumn{2}{l|}{\textsc{en-ru}}       & newstest2018 \\
            \multicolumn{2}{l|}{\textsc{en-tr}}       & newstest2018 \\
            \multicolumn{2}{l|}{\textsc{en-zh}}       & newstest2018 \\
            \multicolumn{2}{l|}{\textsc{de-fr}}       & euelections\_dev2019 \\
            \bottomrule
            \end{tabular}
            \caption{The WMT evaluation dataset used for each language pair.}
            \label{tab:evaluation-datasets}
        \end{center}
    \end{table}
}

\newcommand{\insertDatasetInformationTable}{
    \begin{table}[t]
        \begin{center}
            \resizebox{1\linewidth}{!}{
            \begin{tabular}[b]{cc|r|r|c|c|}
            \toprule
             & & \textbf{\# seen}  & \textbf{\# available} & \textbf{\# epochs} & \textbf{\% budget} \\
            \midrule
            \multicolumn{2}{l|}{\textsc{en-cs}}       & 3,466,692 & 51,136,198 & 0.06 & 10.7 \\
            \multicolumn{2}{l|}{\textsc{en-de}}       & 2,678,808 & 3,054,632  & 0.88 & 8.3 \\
            \multicolumn{2}{l|}{\textsc{en-fi}}       & 3,466,692 & 6,457,071  & 0.54 & 10.7 \\
            \multicolumn{2}{l|}{\textsc{en-gu}}       & 1,260,615 & 137,905    & 9.14 & 3.9 \\
            \multicolumn{2}{l|}{\textsc{en-kk}}       & 1,181,827 & 158,067    & 7.47 & 3.7 \\
            \multicolumn{2}{l|}{\textsc{en-lt}}       & 3,624,269 & 2,283,272  & 1.59 & 11.2 \\
            \multicolumn{2}{l|}{\textsc{en-ru}}       & 5,042,462 & 11,391,126 & 0.44 & 15.6 \\
            \multicolumn{2}{l|}{\textsc{en-tr}}       & 1,575,769 & 207,678    & 7.58 & 4.9 \\
            \multicolumn{2}{l|}{\textsc{en-zh}}       & 5,846,104 & 14,549,833 & 0.40 & 18.1 \\
            \multicolumn{2}{l|}{\textsc{de-fr}}       & 4,097,000 & 1,980,332  & 2.06 & 12.7 \\
            \midrule
            \multicolumn{2}{l|}{\textsc{Total}}       & 32,240,238 & 91,356,114  & - &  100 \\
            
            \bottomrule
            \end{tabular}
            }
            \caption{Training dataset statistics for our multilingual NMT experiments. \textbf{\# seen} is the total number of segments seen during training. \textbf{\# available} is the number of unique segments available in the parallel training datasets. \textbf{\# epochs} is the number of passes made over the available training data -- when this is $<1$, the available training data was only partially seen. \textbf{\% budget} is the percentage of the training budget allocated to this pair of tasks.}
            \label{tab:dataset-information}
        \end{center}
    \end{table}
}

\newcommand{\insertParallelTaskTable}{
    \begin{table}[t]
        \begin{center}
            \resizebox{1\linewidth}{!}{
            \begin{tabular}[b]{cc|c|c|c|c|}
            \toprule
             & & \textbf{\textsc{Prepend}} & \textbf{\textsc{Emb}} & \textbf{\textsc{Dec}} & \textbf{\textsc{Attn}} \\
            \midrule
            \multicolumn{2}{l|}{\textsc{cs-en}}       & \bleu{20.192} & \bleu{20.234} & \bestbleu{20.945} & \bleu{20.894} \\
            \multicolumn{2}{l|}{\textsc{en-cs}}       & \bleu{12.365} & \bleu{12.703} & \bestbleu{13.680} & \bleu{13.304} \\
            \multicolumn{2}{l|}{\textsc{de-en}}       & \bleu{26.204} & \bleu{26.064} & \bestbleu{27.430} & \bleu{27.078} \\
            \multicolumn{2}{l|}{\textsc{en-de}}       & \bleu{23.201} & \bleu{23.406} & \bestbleu{25.706} & \bleu{25.240} \\
            \multicolumn{2}{l|}{\textsc{fi-en}}       & \bleu{13.713} & \bleu{13.542} & \bestbleu{14.355} & \bleu{14.159} \\
            \multicolumn{2}{l|}{\textsc{en-fi}}       & \bleu{8.316} & \bleu{8.049} & \bestbleu{9.354} & \bleu{9.173} \\
            \multicolumn{2}{l|}{\textsc{gu-en}}       & \bleu{15.43} & \bleu{15.439} & \bestbleu{15.666} & \bleu{15.403} \\
            \multicolumn{2}{l|}{\textsc{en-gu}}       & \bestbleu{8.098} & \bleu{7.770} & \bleu{5.107} & \bleu{7.256} \\
            \multicolumn{2}{l|}{\textsc{kk-en}}       & \bestbleu{14.42} & \bleu{13.985} & \bleu{14.336} & \bleu{13.880} \\
            \multicolumn{2}{l|}{\textsc{en-kk}}       & \textbf{\bleu{5.557}} & \bleu{5.197} & \bleu{1.851} & \bleu{4.614} \\
            \multicolumn{2}{l|}{\textsc{lt-en}}       & \bleu{18.550} & \bleu{18.865} & \textbf{\bleu{19.345}} & \bleu{19.047} \\
            \multicolumn{2}{l|}{\textsc{en-lt}}       & \bleu{12.754} & \bleu{13.022} & \textbf{\bleu{14.406}} & \bleu{13.726} \\
            \multicolumn{2}{l|}{\textsc{ru-en}}       & \bleu{20.752} & \bleu{20.573} & \bleu{21.323} & \textbf{\bleu{21.344}} \\
            \multicolumn{2}{l|}{\textsc{en-ru}}       & \bleu{15.538} & \bleu{15.930} & \textbf{\bleu{17.029}} & \bleu{16.735} \\
            \multicolumn{2}{l|}{\textsc{tr-en}}       & \bleu{14.81} & \bleu{14.953} & \textbf{\bleu{15.188}} & \bleu{15.074} \\
            \multicolumn{2}{l|}{\textsc{en-tr}}       & \bleu{10.318} & \bleu{10.034} & \bleu{10.931} & \textbf{\bleu{11.332}} \\
            \multicolumn{2}{l|}{\textsc{zh-en}}       & \bleu{13.53} & \bleu{13.705} & \textbf{\bleu{14.069}} & \bleu{13.727} \\
            \multicolumn{2}{l|}{\textsc{en-zh}}       & \bleu{24.2} & \bleu{24.4} & \textbf{\bleu{25.6}} & \bleu{25.4} \\
            \multicolumn{2}{l|}{\textsc{fr-de}}       & \bleu{18.553} & \bleu{18.435} & \textbf{\bleu{19.89}} & \bleu{19.294} \\
            \multicolumn{2}{l|}{\textsc{de-fr}}       & \bleu{21.170} & \bleu{22.114} & \bleu{21.730} & \textbf{\bleu{22.566}} \\
            
            \bottomrule
            \end{tabular}
            }
            \caption{Results for all task pairs in the WMT 2019 news-translation shared task where parallel training data is available.}
            \label{tab:parallel-tasks}
        \end{center}
    \end{table}
}

\newcommand{\insertGlobalBleuTable}{
    \begin{table}[t]
        \begin{center}
            \resizebox{1\linewidth}{!}{
            \begin{tabular}[b]{cc|c|c|c|c|}
            \toprule
             & & \textbf{\textsc{Prepend}} & \textbf{\textsc{Emb}} & \textbf{\textsc{Dec}} & \textbf{\textsc{Attn}} \\
            \midrule
            \multicolumn{2}{l|}{\textsc{Supervised}}       & \bleu{23.382} & \bleu{23.405} & \bleu{23.963} & \textbf{\bleu{24.071}} \\
            \multicolumn{2}{l|}{\textsc{Zero-Shot-TED}}       & \bleu{10.606} & \bleu{7.790} & \bestbleu{12.601} & \bleu{12.364} \\
            \multicolumn{2}{l|}{\textsc{Zero-Shot-Pivot}}       & \bleu{16.866} & \textbf{\bleu{18.061}} & \bleu{13.952} & \bleu{15.143} \\
            \multicolumn{2}{l|}{\textsc{Zero-Shot-Parallel-Pivot}}       & \bleu{12.646} & \bleu{11.256} & \bleu{12.277} & \bestbleu{13.082} \\
            \bottomrule
            \end{tabular}
            }
            \caption{Overall results for supervised and zero-shot tasks. Tokenized BLEU scores are computed by concatenating all of the hypotheses for all translation directions, and computing BLEU with respect to the concatenated references. We use the sentencepiece-segmented hypotheses and references to avoid issues with tokenization of multi-lingual hypotheses and references.}
            \label{tab:global-bleu}
        \end{center}
    \end{table}
}

\newcommand{\insertFailedTasksTable}{
    \begin{table}[t]
        \begin{center}
            \resizebox{1\linewidth}{!}{
            \begin{tabular}[b]{cc|c|c|c|c|}
            \toprule
             & & \textbf{\textsc{Prepend}} & \textbf{\textsc{Emb}} & \textbf{\textsc{Dec}} & \textbf{\textsc{Attn}} \\
            \midrule
            \multicolumn{2}{l|}{\# Failed Pivot Tasks}       & 3 & 31 & 1 & 1 \\
            \bottomrule
            \end{tabular}
            }
            \caption{Out of 110 pivot translation tasks, how many failed the language identification check?}
            \label{tab:failed-pivot-tasks}
        \end{center}
    \end{table}
}

\newcommand{\insertShortZeroshotTable}{
    \begin{table}[t]
        \begin{center}
            \resizebox{1\linewidth}{!}{
            \begin{tabular}[b]{cc|c|c|c|c|}
            \toprule
             & & \textbf{\textsc{Prepend}} & \textbf{\textsc{Emb}} & \textbf{\textsc{Dec}} & \textbf{\textsc{Attn}} \\
            \midrule
            \multicolumn{2}{l|}{\textsc{ru$\rightarrow$cs$\rightarrow$ru}}  & \bleu{20.867} & \textbf{\bleu{23.843}} & \bleu{20.814} & \bleu{20.963} \\
            \multicolumn{2}{l|}{\textsc{ru$\rightarrow$de$\rightarrow$ru}}  & \bleu{14.616} & \bleu{11.872} & \textbf{\bleu{16.365}} & \bleu{15.590} \\
            \multicolumn{2}{l|}{\textsc{ru$\rightarrow$en$\rightarrow$ru}*}  & \bleu{21.674} & \bleu{22.224} & \textbf{\bleu{23.937}} & \bleu{23.180} \\
            \multicolumn{2}{l|}{\textsc{ru$\rightarrow$fi$\rightarrow$ru}}  & \bleu{11.161} & \textbf{\bleu{17.001}} & \bleu{12.059} & \bleu{11.589} \\
            \multicolumn{2}{l|}{\textsc{ru$\rightarrow$fr$\rightarrow$ru}}  & \bleu{13.755} & \textbf{\bleu{15.351}} & \bleu{14.098} & \bleu{15.149} \\
            \multicolumn{2}{l|}{\textsc{ru$\rightarrow$gu$\rightarrow$ru}}  & \textbf{\bleu{10.3}} & \bleu{9.6} & \bleu{3.5} & \bleu{5.1} \\
            \multicolumn{2}{l|}{\textsc{ru$\rightarrow$kk$\rightarrow$ru}}  & \sout{\bleu{5.766}} & \sout{\bleu{19.565}} & \sout{\bleu{1.030}} & \sout{\bleu{2.212}} \\
            \multicolumn{2}{l|}{\textsc{ru$\rightarrow$lt$\rightarrow$ru}}  & \bleu{16.880} & \textbf{\bleu{22.037}} & \bleu{16.536} & \bleu{16.639} \\
            \multicolumn{2}{l|}{\textsc{ru$\rightarrow$tr$\rightarrow$ru}}  & \bleu{7.914} & \textbf{\bleu{10.232}} & \bleu{7.447} & \bleu{7.661} \\
            \multicolumn{2}{l|}{\textsc{ru$\rightarrow$zh$\rightarrow$ru}}  & \bleu{8.843} & \textbf{\bleu{10.474}} & \bleu{9.099} & \bleu{8.483} \\
            \bottomrule
            \end{tabular}
            }
            \caption{Zero-shot translation results for \textsc{ru$\rightarrow$*$\rightarrow$ru}  Note that BLEU scores are computed by translating $\textsc{SRC} \rightarrow \textsc{Pivot} \rightarrow \textsc{\reallywidehat{SRC}}$, and computing the score between $\textsc{SRC}$ and $\textsc{\reallywidehat{SRC}}$. Systems which do not pass the language identification filter are struck-through and removed from global evaluation. Note that parallel training data was available for \textsc{ru$\rightarrow$en}.}
            \label{tab:zero-shot-ru}
        \end{center}
    \end{table}
}

\newcommand{\insertMTMatrixZeroshotTable}{
    \begin{table*}[!htbp]
        \begin{center}
            \scriptsize
            \resizebox{1\linewidth}{!}{
            \begin{tabular}[b]{r|c|c|c|c|c|c|c|c|c|c|c|}
            \cmidrule[1pt]{2-12}
            
            \textsc{Prepend} & \multirow{4}*{\textbf{\textsc{cs}}} & \bleu{15.72}     & \bleu{19.64}     & \bleu{11.36}        & \bleu{11.10}     & \bleu{8.67}     & \bleu{3.55} & \bleu{16.57}    & \bleu{17.41}     & \bleu{7.75}      & \bleu{7.52} \\ 
            \textsc{Emb} &                                         & \bleu{9.30}      & \bleu{19.91}     & \sout{\bleu{24.36}} & \bleu{12.39}     & \bleu{9.28} & \bleu{3.32}     & \sout{\bleu{28.41}} & \bleu{15.80}     & \sout{\bleu{10.16}} & \bleu{7.53} \\ 
            \textsc{Dec} &                                         & \bleu{17.72} & \bleu{21.26}     & \bleu{11.45}        & \bleu{13.32}     & \bleu{3.26}     & \bleu{0.74}     & \bleu{13.95}        & \bleu{17.83} & \bleu{6.72}      & \bleu{7.51} \\ 
            \textsc{Attn} &                                        & \bleu{17.45}     & \bleu{21.55} & \bleu{11.60}    & \bleu{13.81} & \bleu{4.49}     & \bleu{1.84}     & \bleu{14.36}        & \bleu{17.42}     & \bleu{7.21}      & \bleu{7.60} \\ 
            \cmidrule{2-12}
            \textsc{Prepend} & \bleu{22.25}    & \multirow{4}*{\textbf{\textsc{de}}}  & \bleu{27.05}     & \bleu{16.37}    & \bleu{25.32}     & \bleu{11.20}    & \bleu{5.32}     & \bleu{18.29}    & \bleu{18.13}     & \bleu{12.39} & \bleu{12.34} \\ 
            \textsc{Emb} & \sout{\bleu{41.51}} &                                      & \bleu{27.62}     & \sout{\bleu{38.00}} & \bleu{25.08}     & \sout{\bleu{19.42}} & \bleu{8.47} & \sout{\bleu{40.67}} & \bleu{23.57} & \sout{\bleu{29.89}} & \sout{\bleu{19.62}} \\ 
            \textsc{Dec} & \bleu{23.35} &                                         & \bleu{29.81} & \bleu{15.64}        & \bleu{25.84}     & \bleu{3.99}         & \bleu{0.93}     & \bleu{16.22}        & \bleu{19.14}     & \bleu{11.23} & \bleu{12.19} \\ 
            \textsc{Attn} & \bleu{22.79} &                                            & \bleu{29.02}     & \bleu{15.91}        & \bleu{26.33} & \bleu{6.40}         & \bleu{2.74}     & \bleu{17.05}        & \bleu{17.99}     & \bleu{11.19} & \bleu{12.06} \\ 
            \cmidrule{2-12}
            \textsc{Prepend} & \bleu{35.41}            & \bleu{37.05} & \multirow{4}*{\textbf{\textsc{en}}} & \bleu{24.62}     & \bleu{34.46} & \bleu{22.13} & \bleu{9.26} & \bleu{29.18} & \bleu{32.94} & \bleu{23.29} & \bleu{25.61} \\ 
            \textsc{Emb}     & \bleu{36.53} & \bleu{37.44} &                                                & \bleu{25.89} & \sout{\bleu{35.00}} & \bleu{21.49} & \bleu{9.26} & \bleu{30.37} & \bleu{33.85} & \bleu{24.24} & \bleu{26.57} \\ 
            \textsc{Dec}     & \bleu{35.82} & \bleu{37.49} &                                                & \bleu{25.76}     & \bleu{32.62} & \bleu{10.11} & \bleu{1.58} & \bleu{29.71} & \bleu{33.23} & \bleu{22.76} & \bleu{26.34} \\ 
            \textsc{Attn}    & \bleu{36.89} & \bleu{36.60} &                                                & \bleu{25.88}     & \bleu{34.37} & \bleu{15.66} & \bleu{6.20} & \bleu{30.33} & \bleu{33.86} & \bleu{23.63} & \bleu{26.80} \\ 
            \cmidrule{2-12}
            \textsc{Prepend} & \bleu{12.05} & \bleu{10.98} & \bleu{14.33} & \multirow{4}*{\textbf{\textsc{fi}}} & \bleu{7.15} & \bleu{5.62} & \bleu{2.55} & \sout{\bleu{13.10}} & \bleu{9.16}   & \bleu{6.59}  & \bleu{6.24} \\ 
            \textsc{Emb} & \sout{\bleu{19.86}} & \sout{\bleu{7.87}}  & \bleu{14.65} &                                         & \bleu{8.17} & \bleu{6.20} & \bleu{4.22} & \sout{\bleu{23.76}} & \bleu{11.14}  & \sout{\bleu{12.00}} & \sout{\bleu{6.76}} \\ 
            \textsc{Dec} & \bleu{11.23} & \bleu{11.69} & \bleu{15.41} &                                         & \bleu{9.77} & \bleu{3.02} & \bleu{0.54} & \bleu{10.74} & \bleu{9.85}   & \bleu{6.15}  & \bleu{5.78} \\ 
            \textsc{Attn} & \bleu{12.15} & \bleu{11.48} & \bleu{15.01} &                                        & \bleu{10.00} & \bleu{4.18} & \bleu{1.70} & \bleu{10.87} & \bleu{9.77}   & \bleu{6.47}  & \bleu{5.79} \\ 
            \cmidrule{2-12}
            \textsc{Prepend} & \bleu{25.56} & \bleu{32.65} & \bleu{31.87} & \bleu{17.83} & \multirow{4}*{\textbf{\textsc{fr}}} & \bleu{17.10} & \bleu{7.79} & \bleu{20.47} & \bleu{22.88} & \bleu{16.04} & \bleu{15.27} \\ 
            \textsc{Emb} & \bleu{25.99} & \bleu{32.02} & \sout{\bleu{36.53}} & \bleu{20.42} &                                         & \bleu{12.41} & \bleu{5.02} & \sout{\bleu{24.53}} & \bleu{22.49} & \bleu{15.58} & \bleu{14.10} \\ 
            \textsc{Dec} & \bleu{25.68} & \bleu{32.90} & \bleu{33.91} & \bleu{18.39} &                                         & \bleu{5.86} & \bleu{1.51} & \bleu{20.24} & \bleu{23.61} & \bleu{14.44} & \bleu{15.07} \\ 
            \textsc{Attn} & \bleu{26.04} & \bleu{33.16} & \bleu{34.34} & \bleu{19.54} &                                        & \bleu{8.51} & \bleu{4.54} & \bleu{20.99} & \bleu{24.56} & \bleu{15.49} & \bleu{15.36} \\ 
            \cmidrule{2-12}
            \textsc{Prepend} & \bleu{5.14} & \bleu{5.72} & \bleu{8.17} & \bleu{4.17} & \bleu{3.98} & \multirow{4}*{\textbf{\textsc{gu}}} & \bleu{2.01} &  \bleu{5.29} & \bleu{5.20} & \bleu{3.50} & \bleu{3.94} \\ 
            \textsc{Emb}     & \sout{\bleu{5.30}} & \bleu{4.64} & \bleu{7.62} & \bleu{5.39} & \bleu{4.37} &                                     & \bleu{1.55} &  \sout{\bleu{6.22}} & \bleu{3.60} & \bleu{4.20} & \bleu{3.67} \\ 
            \textsc{Dec}     & \bleu{3.54} & \bleu{3.50} & \bleu{5.48} & \bleu{3.24} & \bleu{3.23} &                                     & \bleu{0.51} &  \bleu{3.56} & \bleu{3.39} & \bleu{2.86} & \bleu{2.65} \\ 
            \textsc{Attn}    & \bleu{4.88} & \bleu{5.05} & \bleu{7.56} & \bleu{4.43} & \bleu{4.52} &                                     & \bleu{1.04} & \bleu{4.60} & \bleu{4.53} & \bleu{4.04} & \bleu{4.00} \\ 
            \cmidrule{2-12}
            \textsc{Prepend} & \bleu{4.86} & \bleu{4.77} & \bleu{7.05} & \bleu{3.40} & \bleu{2.83} & \bleu{4.29} & \multirow{4}*{\textbf{\textsc{kk}}}  & \bleu{4.61} & \bleu{4.90} & \bleu{4.43} & \bleu{3.88} \\ 
            \textsc{Emb}     & \bleu{3.36} & \bleu{3.40} & \bleu{6.36} & \bleu{3.82} & \bleu{3.50} & \sout{\bleu{1.87}} &                                      & \bleu{3.93} & \sout{\bleu{3.62}} & \bleu{4.48} & \bleu{2.40} \\ 
            \textsc{Dec}     & \bleu{1.65} & \bleu{1.76} & \bleu{2.18} & \bleu{1.55} & \bleu{1.47} & \bleu{1.07} &                                      & \bleu{1.53} & \bleu{1.65} & \bleu{1.74} & \bleu{1.19} \\ 
            \textsc{Attn}    & \bleu{3.89} & \bleu{3.96} & \bleu{5.34} & \bleu{3.22} & \bleu{3.19} & \bleu{2.26} &                                      & \bleu{3.61} & \bleu{3.89} & \bleu{4.31} & \bleu{3.19} \\ 
            \cmidrule{2-12}
            \textsc{Prepend} & \bleu{18.84} & \bleu{14.48} & \bleu{17.92} & \bleu{13.81} & \bleu{10.37} & \bleu{9.88} & \bleu{4.67} & \multirow{4}*{\textbf{\textsc{lt}}} & \bleu{16.82} & \bleu{9.99}  & \bleu{8.98} \\ 
            \textsc{Emb}     & \sout{\bleu{30.52}} & \sout{\bleu{11.99}} & \bleu{18.33} & \sout{\bleu{30.70}} & \bleu{10.82} & \bleu{10.60} & \sout{\bleu{6.75}} &                                     & \bleu{18.21} & \sout{\bleu{19.71}} & \bleu{13.71} \\ 
            \textsc{Dec}     & \bleu{16.74} & \bleu{13.56} & \bleu{18.87} & \bleu{12.80} & \bleu{11.60} & \bleu{3.66} & \bleu{0.93} &                                     & \bleu{16.10} & \bleu{8.03}  & \bleu{8.54} \\ 
            \textsc{Attn}    & \bleu{16.95} & \bleu{13.86} & \bleu{18.75} & \bleu{12.54} & \bleu{12.36} & \bleu{5.46} & \bleu{1.89} &                                     & \bleu{15.70} & \bleu{8.72}  & \bleu{9.16} \\ 
            \cmidrule{2-12} 
            
            \textsc{Prepend} & \bleu{20.87} & \bleu{14.62} & \bleu{21.67} & \bleu{11.16} & \bleu{13.76} & \bleu{10.26} & \sout{\bleu{5.77}}  & \bleu{16.88} & \multirow{4}*{\textbf{\textsc{ru}}} & \bleu{7.91}  & \bleu{8.84} \\ 
            \textsc{Emb}     & \bleu{23.84} & \bleu{11.87} & \bleu{22.22} & \bleu{17.00} & \bleu{15.35} & \bleu{9.56} & \sout{\bleu{19.57}} & \sout{\bleu{22.04}} &                                     & \bleu{10.23} & \bleu{10.47} \\ 
            \textsc{Dec}     & \bleu{20.81} & \bleu{16.37} & \bleu{23.94} & \bleu{12.06} & \bleu{14.10} & \bleu{3.54} & \sout{\bleu{1.03}}  & \bleu{16.54} &                                     & \bleu{7.45}  & \bleu{9.10} \\ 
            \textsc{Attn}    & \bleu{20.96} & \bleu{15.59} & \bleu{23.18} & \bleu{11.59} & \bleu{15.15} & \bleu{5.05} & \sout{\bleu{2.21}}  & \bleu{16.64} &                                     & \bleu{7.66}  & \bleu{8.48} \\ 
            \cmidrule{2-12}
            
            \textsc{Prepend} & \bleu{9.11}  & \bleu{8.17} & \bleu{13.33} & \bleu{7.42}  & \bleu{7.35} & \bleu{8.77} & \bleu{5.60} & \bleu{8.86}  & \bleu{7.11} &  \multirow{4}*{\textbf{\textsc{tr}}} & \bleu{6.40} \\ 
            \textsc{Emb}     & \sout{\bleu{12.66}} & \bleu{7.19} & \bleu{12.64} & \sout{\bleu{14.58}} & \bleu{7.48} & \bleu{6.65} & \bleu{4.10} & \sout{\bleu{17.28}} & \bleu{6.62} &                                      & \bleu{6.36} \\ 
            \textsc{Dec}     & \bleu{7.15}  & \bleu{7.62} & \bleu{13.05} & \bleu{6.47}  & \bleu{6.83} & \bleu{2.50} & \bleu{0.65} & \bleu{6.44}  & \bleu{5.63} &                                      & \bleu{5.28} \\ 
            \textsc{Attn}    & \bleu{7.33}  & \bleu{8.13} & \bleu{13.41} & \bleu{6.63}  & \bleu{7.27} & \bleu{3.92} & \bleu{1.89} & \bleu{6.73}  & \bleu{5.56} &                                      & \bleu{5.26} \\ 
            \cmidrule{2-12}
            
            \textsc{Prepend} & \bleu{20.4} & \bleu{19.6} & \bleu{29.0} & \bleu{17.1} & \bleu{17.4} & \sout{\bleu{18.2}} & \bleu{8.4} & \bleu{20.2} & \bleu{19.5} & \bleu{17.4} & \multirow{4}*{\textbf{\textsc{zh}}} \\ 
            \textsc{Emb}     & \bleu{20.1} & \bleu{16.9} & \bleu{29.4} & \bleu{19.6} & \bleu{17.8} & \bleu{11.9} & \bleu{6.6} & \bleu{22.8} & \bleu{18.3} & \bleu{16.7} & \\ 
            \textsc{Dec}     & \bleu{19.2} & \bleu{19.4} & \bleu{30.2} & \bleu{16.6} & \bleu{17.6} & \bleu{7.2} & \bleu{2.2} & \bleu{19.5} & \bleu{20.1} & \bleu{16.3} & \\ 
            \textsc{Attn}    & \bleu{19.8} & \bleu{20.4} & \bleu{30.0} & \bleu{16.7} & \bleu{18.2} & \bleu{11.0} & \bleu{5.0} & \bleu{18.6} & \bleu{19.4} & \bleu{17.1} & \\ 
            \cmidrule[1pt]{2-12}
            \end{tabular}
            }
            \caption{\textsc{Zero-Shot Pivot} translation results in all directions, for all models. Rows indicate source language, columns indicate pivot language.  For example, cell $(1, 2)$ contains the results for  \textsc{cs$\rightarrow$de$\rightarrow$cs}. Runs which did not pass the language identification filter are struck-through. The MT-matrix (\url{http://matrix.statmt.org/matrix}) was the inspiration for this rendering. Note this table shows results for all translation directions, but we only use the zero-shot pairs for evaluation.}
            \label{tab:pivot-translation-full-results}
        \end{center}
    \end{table*}
}

\newcommand{\insertParallelPivotMTMatrixTable}{
    \begin{table*}[!htbp]
        \begin{center}
            \scriptsize
            \resizebox{1\linewidth}{!}{
            \begin{tabular}[b]{r|c|c|c|c|c|c|c|c|c|c|}
            \cmidrule[1pt]{2-11}
        \textsc{Prepend} &  \multirow{4}*{\textbf{\textsc{cs}}} &  \bleu{14.414874476078497} &  \bleu{6.088300056286523} &  \bleu{14.935391794352618} &  \bleu{3.635169955693632} &  \bleu{3.658600207726975} &  \bleu{9.421033501890738} &  \bleu{10.471132066757772} &  \bleu{6.3999319949550415} &  \bleu{18.1} \\
\textsc{Emb} &  \bleu{} &  \bleu{11.520446690250273} &  \bleu{3.7374944055431} &  \bleu{17.338392140179163} &  \bleu{2.6278738590608524} &  \bleu{2.5756987721102145} &  \bleu{6.172204545169574} &  \bleu{9.076473467923657} &  \bleu{4.219306814316005} &  \bleu{14.3} \\
\textsc{Dec} &  \bleu{} &  \bleu{15.632544171663184} &  \bleu{6.3968945565484825} &  \bleu{15.454741952185957} &  \bleu{3.1090181362912768} &  \bleu{1.6083187243599644} &  \bleu{10.234462099948114} &  \bleu{11.762473035584188} &  \bleu{6.860612723941004} &  \bleu{19.7} \\
\textsc{Attn} &  \bleu{} &  \bleu{16.114497912009874} &  \bleu{6.55528092685287} &  \bleu{15.348641271057106} &  \bleu{4.2985776579752555} &  \bleu{3.2626246929086093} &  \bleu{9.743903318544128} &  \bleu{11.572168996493767} &  \bleu{7.124717894702883} &  \bleu{20.1} \\
\cmidrule[1pt]{2-11}
\textsc{Prepend} &  \bleu{8.529012978111979} &  \multirow{4}*{\textbf{\textsc{de}}} &  \bleu{6.204265103992004} &  \bleu{} &  \bleu{4.61521680370171} &  \bleu{3.8460435399949926} &  \bleu{8.588704000658291} &  \bleu{9.837263426544263} &  \bleu{6.3611289628876255} &  \bleu{18.4} \\
\textsc{Emb} &  \bleu{3.8995217934150683} &  \bleu{} &  \bleu{3.2807713048584577} &  \bleu{} &  \bleu{2.51814079373133} &  \bleu{2.535293037387257} &  \bleu{4.651232615675921} &  \bleu{7.107404524144222} &  \bleu{3.632683733352523} &  \bleu{14.4} \\
\textsc{Dec} &  \bleu{9.97564860949643} &  \bleu{} &  \bleu{7.272112396298628} &  \bleu{} &  \bleu{3.3237017390733214} &  \bleu{1.5409987692762148} &  \bleu{9.118239802147878} &  \bleu{10.902695232958196} &  \bleu{7.357109338928602} &  \bleu{20.4} \\
\textsc{Attn} &  \bleu{9.81714787032151} &  \bleu{} &  \bleu{7.309903125052365} &  \bleu{} &  \bleu{4.462257571072914} &  \bleu{3.343272829533058} &  \bleu{9.174772509031467} &  \bleu{10.874952625972854} &  \bleu{7.460488213823105} &  \bleu{20.6} \\
\cmidrule[1pt]{2-11}
\textsc{Prepend} &  \bleu{6.603005125045748} &  \bleu{10.53518550789982} &  \multirow{4}*{\textbf{\textsc{fi}}} &  \bleu{11.483609152547563} &  \bleu{3.3933835849538654} &  \bleu{2.7946723922154755} &  \bleu{7.662530326333799} &  \bleu{7.572185485460277} &  \bleu{5.121476903438894} &  \bleu{15.0} \\
\textsc{Emb} &  \bleu{4.04168542081464} &  \bleu{8.629108904386111} &  \bleu{} &  \bleu{16.277060931629244} &  \bleu{2.664907617310452} &  \bleu{1.7868754839611432} &  \bleu{5.689423928328953} &  \bleu{5.7941875946364565} &  \bleu{3.4361698112338357} &  \bleu{12.5} \\
\textsc{Dec} &  \bleu{7.543628517658198} &  \bleu{12.32351196416432} &  \bleu{} &  \bleu{12.024796538940151} &  \bleu{3.136648302610435} &  \bleu{1.4620543229620053} &  \bleu{8.319030530455747} &  \bleu{8.54233570679011} &  \bleu{5.398713804085403} &  \bleu{16.8} \\
\textsc{Attn} &  \bleu{7.440593966391604} &  \bleu{12.102395510828641} &  \bleu{} &  \bleu{12.345944260775198} &  \bleu{3.8094835966654843} &  \bleu{2.915752521723301} &  \bleu{8.10970343332754} &  \bleu{7.959902373162625} &  \bleu{5.457070325591112} &  \bleu{16.7} \\
\cmidrule[1pt]{2-11}
\textsc{Prepend} &  \bleu{} &  \bleu{16.492407179331877} &  \bleu{4.752291842207557} &  \multirow{4}*{\textbf{\textsc{fr}}} &  \bleu{3.471726686742642} &  \bleu{2.824734353118774} &  \bleu{7.372021095079622} &  \bleu{9.251598700545761} &  \bleu{5.9764242193331105} &  \bleu{17.2} \\
\textsc{Emb} &  \bleu{} &  \bleu{17.053799386199714} &  \bleu{5.266444633250403} &  \bleu{} &  \bleu{3.9478626923037066} &  \bleu{3.1097108782190865} &  \bleu{8.31906538377209} &  \bleu{10.971688625246923} &  \bleu{6.81410747370256} &  \bleu{18.5} \\
\textsc{Dec} &  \bleu{} &  \bleu{16.121694982857697} &  \bleu{6.347529574193385} &  \bleu{} &  \bleu{2.8634608599456746} &  \bleu{1.4227110618901895} &  \bleu{8.06349204697799} &  \bleu{10.590159179325584} &  \bleu{6.505893156035486} &  \bleu{18.7} \\
\textsc{Attn} &  \bleu{} &  \bleu{16.330666025143163} &  \bleu{6.236562303120987} &  \bleu{} &  \bleu{4.877199895522516} &  \bleu{2.972976714993755} &  \bleu{8.511123673596467} &  \bleu{10.544058867953233} &  \bleu{6.795706716422605} &  \bleu{19.4} \\
\cmidrule[1pt]{2-11}
\textsc{Prepend} &  \bleu{4.233605321313369} &  \bleu{6.926930321812668} &  \bleu{3.1672424949699036} &  \bleu{7.8899261284051025} &  \multirow{4}*{\textbf{\textsc{gu}}} &  \bleu{2.4090420213539896} &  \bleu{5.090794046286559} &  \bleu{5.455129103849} &  \bleu{3.3575183831646833} &  \bleu{11.5} \\
\textsc{Emb} &  \bleu{2.4944714580661964} &  \bleu{5.561736971923027} &  \bleu{2.2713815179715984} &  \bleu{9.026466247162249} &  \bleu{} &  \bleu{1.8731824270950734} &  \bleu{3.4723313202664845} &  \bleu{3.522197450529081} &  \bleu{2.43240063232057} &  \bleu{8.7} \\
\textsc{Dec} &  \bleu{2.4308490284580584} &  \bleu{3.9621386262723854} &  \bleu{2.0109491573914964} &  \bleu{3.975855379856164} &  \bleu{} &  \bleu{1.0348947348469921} &  \bleu{3.0859579487533626} &  \bleu{3.106690936579163} &  \bleu{2.2277297577035595} &  \bleu{7.6} \\
\textsc{Attn} &  \bleu{3.6526618380210465} &  \bleu{5.726423710912327} &  \bleu{2.8942659848750423} &  \bleu{5.798558288076469} &  \bleu{} &  \bleu{2.0554700405181916} &  \bleu{4.28355460818796} &  \bleu{4.058425462342932} &  \bleu{3.0528711409360882} &  \bleu{10.5} \\
\cmidrule[1pt]{2-11}
\textsc{Prepend} &  \bleu{2.082328088855985} &  \bleu{2.4068335274259303} &  \bleu{1.6684808465183012} &  \bleu{4.878016228117733} &  \bleu{1.3041879753825305} &  \multirow{4}*{\textbf{\textsc{kk}}} &  \bleu{2.94711941558847} &  \bleu{2.726690632145856} &  \bleu{2.122602396039762} &  \bleu{7.9} \\
\textsc{Emb} &  \bleu{1.1763002128922255} &  \bleu{2.0208490668073713} &  \bleu{1.325865758127423} &  \bleu{5.083644512158225} &  \bleu{1.0598329388607357} &  \bleu{} &  \bleu{1.987296375785886} &  \bleu{1.9243953579852604} &  \bleu{1.8102655738093685} &  \bleu{5.7} \\
\textsc{Dec} &  \bleu{0.7436673225389391} &  \bleu{0.6416143113357395} &  \bleu{0.5544210341884183} &  \bleu{1.6724637085568734} &  \bleu{0.3539728930507486} &  \bleu{} &  \bleu{0.8124260127294499} &  \bleu{0.8638684544588268} &  \bleu{0.5463354402301144} &  \bleu{2.9} \\
\textsc{Attn} &  \bleu{1.6227562900925445} &  \bleu{1.8530222116825237} &  \bleu{1.7505075544128408} &  \bleu{3.5700356560924456} &  \bleu{1.0221570511625844} &  \bleu{} &  \bleu{1.8231691893480668} &  \bleu{2.042293802158492} &  \bleu{1.6597908602118912} &  \bleu{6.4} \\
\cmidrule[1pt]{2-11}
\textsc{Prepend} &  \bleu{7.547327748583423} &  \bleu{10.988432243838474} &  \bleu{5.8362958782991} &  \bleu{13.16062306841417} &  \bleu{3.908913549412104} &  \bleu{3.4363155554007987} &  \multirow{4}*{\textbf{\textsc{lt}}} &  \bleu{9.167396158602784} &  \bleu{5.432153976386472} &  \bleu{17.6} \\
\textsc{Emb} &  \bleu{4.605523095176305} &  \bleu{9.291767262879674} &  \bleu{3.6713324147199025} &  \bleu{16.9684884256659} &  \bleu{2.4212969051626545} &  \bleu{2.2083290253546677} &  \bleu{} &  \bleu{7.272494853071928} &  \bleu{3.801076861876731} &  \bleu{14.0} \\
\textsc{Dec} &  \bleu{8.277131344355158} &  \bleu{11.896042941835896} &  \bleu{6.236987447643494} &  \bleu{12.891424912416221} &  \bleu{3.0406385156593476} &  \bleu{1.5592147934711684} &  \bleu{} &  \bleu{10.596010897391071} &  \bleu{6.088103747599187} &  \bleu{18.5} \\
\textsc{Attn} &  \bleu{8.322656730311898} &  \bleu{12.012136505207325} &  \bleu{6.547986484687432} &  \bleu{13.156076259113991} &  \bleu{3.923859517510157} &  \bleu{3.164386674912161} &  \bleu{} &  \bleu{9.857629591881823} &  \bleu{6.238473651431638} &  \bleu{18.7} \\
\cmidrule[1pt]{2-11}
\textsc{Prepend} &  \bleu{8.186783472072133} &  \bleu{12.783884671840948} &  \bleu{5.098822863615995} &  \bleu{13.058385816703934} &  \bleu{3.8413647805786715} &  \bleu{3.990072060561763} &  \bleu{9.334562514192513} &  \multirow{4}*{\textbf{\textsc{ru}}} &  \bleu{4.944812765441785} &  \bleu{16.9} \\
\textsc{Emb} &  \bleu{6.4233248437560855} &  \bleu{11.08600700598098} &  \bleu{2.9916291664675603} &  \bleu{11.94205978536333} &  \bleu{2.8910641466851654} &  \bleu{2.7131305397279246} &  \bleu{6.384892633381313} &  \bleu{} &  \bleu{3.657126338992887} &  \bleu{14.9} \\
\textsc{Dec} &  \bleu{9.24099378730909} &  \bleu{13.639086171253233} &  \bleu{5.7290784631332965} &  \bleu{14.0037354162028} &  \bleu{2.6163081042233727} &  \bleu{1.5522270119264667} &  \bleu{9.910178858714035} &  \bleu{} &  \bleu{5.935797388476217} &  \bleu{19.4} \\
\textsc{Attn} &  \bleu{8.745562001518191} &  \bleu{13.752739145507157} &  \bleu{5.586900898372925} &  \bleu{13.760781168520992} &  \bleu{4.133827826188064} &  \bleu{3.636569533175258} &  \bleu{9.639958918862769} &  \bleu{} &  \bleu{5.920511455681561} &  \bleu{19.2} \\
\cmidrule[1pt]{2-11}
\textsc{Prepend} &  \bleu{5.655714658405949} &  \bleu{8.304829759936329} &  \bleu{4.2218282933830205} &  \bleu{9.814406734225692} &  \bleu{2.3005566645737563} &  \bleu{3.1287335148649906} &  \bleu{6.662391738877283} &  \bleu{6.601636812145794} &  \multirow{4}*{\textbf{\textsc{tr}}} &  \bleu{13.9} \\
\textsc{Emb} &  \bleu{3.163366734167544} &  \bleu{6.4740781246801395} &  \bleu{3.3235676661420275} &  \bleu{13.666011972275493} &  \bleu{1.89634067930077} &  \bleu{2.4497240601517283} &  \bleu{4.665519514179498} &  \bleu{4.314405501303736} &  \bleu{} &  \bleu{11.1} \\
\textsc{Dec} &  \bleu{5.18481019449706} &  \bleu{9.060894250548909} &  \bleu{4.3400688250525725} &  \bleu{9.687977450768525} &  \bleu{2.5542298857087657} &  \bleu{1.6062519804367457} &  \bleu{6.2729709810155505} &  \bleu{5.919588021825274} &  \bleu{} &  \bleu{15.5} \\
\textsc{Attn} &  \bleu{5.1983304489419915} &  \bleu{8.856113699978494} &  \bleu{4.676246100359614} &  \bleu{10.365935784865913} &  \bleu{3.397218151169341} &  \bleu{3.202720158623593} &  \bleu{6.0416664907317} &  \bleu{6.018216234009529} &  \bleu{} &  \bleu{16.2} \\
\cmidrule[1pt]{2-11}
\textsc{Prepend} &  \bleu{4.549090162049668} &  \bleu{7.486307931729135} &  \bleu{3.7707144462387747} &  \bleu{9.479953037856694} &  \bleu{2.638014194315829} &  \bleu{2.6725260900329153} &  \bleu{6.211728742744226} &  \bleu{5.955644043356087} &  \bleu{3.521520799699985} &  \multirow{4}*{\textbf{\textsc{zh}}} \\
\textsc{Emb} &  \bleu{3.6661922988626747} &  \bleu{6.8591047818382105} &  \bleu{2.6744295526355644} &  \bleu{10.666566622674795} &  \bleu{2.515771928963229} &  \bleu{2.0052599468654386} &  \bleu{5.263448807865764} &  \bleu{5.7473890572617075} &  \bleu{3.2043111435737517} &  \bleu{} \\
\textsc{Dec} &  \bleu{5.265268172943661} &  \bleu{7.902216530771674} &  \bleu{4.0325132073787975} &  \bleu{10.460187521640332} &  \bleu{2.0983554967526516} &  \bleu{1.1960579559164892} &  \bleu{6.16545982400196} &  \bleu{7.134309519088071} &  \bleu{4.860487829901768} &  \bleu{} \\
\textsc{Attn} &  \bleu{5.2121087234949925} &  \bleu{7.661761530100979} &  \bleu{4.177677962192277} &  \bleu{11.000200287921622} &  \bleu{3.676491942279176} &  \bleu{2.401345432800162} &  \bleu{6.579181275629187} &  \bleu{6.420570903718975} &  \bleu{4.437867154174928} &  \bleu{} \\
\cmidrule[1pt]{2-11}

        \end{tabular}
        }
        \caption{\textsc{Zero-Shot Parallel Pivot} translation results in all directions, for all models. Rows indicate source language, columns indicate pivot language. Note \textsc{en} is omitted from this table because it is used as source language when translating into the pivot language for all target tasks except \textsc{fr}.}
        \label{tab:parallel-pivot-translation-full-results}
    \end{center}
\end{table*}
}

\newcommand*{\bleu}[1]{\num[detect-weight=true,detect-family=true,round-mode=places,round-precision=1]{#1}}
\newcommand*{\bestbleu}[1]{\textbf{\bleu{#1}}}

\newcommand{\chris}[1]{}
\newcommand{\john}[1]{}
\newcommand{\demian}[1]{}

\title{Evaluating the Supervised and Zero-shot Performance of Multi-lingual Translation Models}

\author{Chris Hokamp \and John Glover \and  Demian Gholipour \\
        Aylien Ltd. \\ Dublin, Ireland \\ \texttt{<first-name>@aylien.com} }

\date{}

\begin{document}
\maketitle
\begin{abstract}
We study several methods for full or partial sharing of the decoder parameters of multilingual NMT models. We evaluate both fully supervised and zero-shot translation performance in 110 unique translation directions using only the WMT 2019 shared task parallel datasets for training. We use additional test sets and re-purpose evaluation methods recently used for unsupervised MT in order to evaluate zero-shot translation performance for language pairs where no gold-standard parallel data is available. To our knowledge, this is the largest evaluation of multi-lingual translation yet conducted in terms of the total size of the training data we use, and in terms of the diversity of zero-shot translation pairs we evaluate. We conduct an in-depth evaluation of the translation performance of different models, highlighting the trade-offs between methods of sharing decoder parameters. We find that models which have task-specific decoder parameters outperform models where decoder parameters are fully shared across all tasks. 

\end{abstract}

\section{Introduction}

Multi-lingual translation models, which can map from multiple source languages into multiple target languages, have recently received significant attention because of the potential for positive transfer between high- and low-resource language pairs, and because of possible efficiency gains enabled by translation models which share parameters across many languages \cite{dong-etal-2015-multi,ha-multilingual-2016,firat-etal-2016-multi,johnson-google-2016,blackwood-etal-2018-multilingual,sachan-neubig-2018-parameter,aharoni-etal-2019-massively}. Multi-lingual models that share parameters across languages can also perform zero-shot translation, translating between language pairs for which no parallel training data is available \cite{gnmt2016,ha-multilingual-2016,johnson-google-2016}.

Although multi-task models have recently been shown to achieve positive transfer for some combinations of NLP tasks, in the context of MT, multi-lingual models do not universally outperform models trained to translate in a single direction when sufficient training data is available. However, the ability to do zero-shot translation may be of practical importance in many cases, as parallel training data is not available for most language pairs \cite{gnmt2016,johnson-google-2016,aharoni-etal-2019-massively}. Therefore, small decreases in the performance of supervised pairs may be admissible if the corresponding gain in zero-shot performance is large.  In addition, zero-shot translation can be used to generate synthetic training data for low- or zero- resource language pairs, making it a practical alternative to the bootstrapping by back-translation approach that has recently been used to build completely unsupervised MT systems \cite{firat-etal-2016-multi,artetxe2018iclr,lampleCDR18,lample2018phrase}. Therefore, understanding the trade-offs between different methods of constructing multi-lingual MT systems is still an important line of research.

Deep sequence-to-sequence models have become the established state-of-the-art for machine translation.
The dominant paradigm continues to be models divided into roughly three high-level components:
\textit{embeddings}, which map discrete tokens into real-valued vectors, \textit{encoders}, which map sequences of vectors into an intermediate representation, and \textit{decoders}, which use the representation from an encoder, combined with a dynamic representation of the current state, and output a sequence of tokens in the target language conditioned upon the encoder's representation of the input. For multi-lingual systems, any combination of embedding, encoder and/or decoder parameters can potentially be shared by groups of tasks, or duplicated and kept private for each task.

\begin{figure}[!ht]  
\centering
\includegraphics[width=0.45\textwidth]{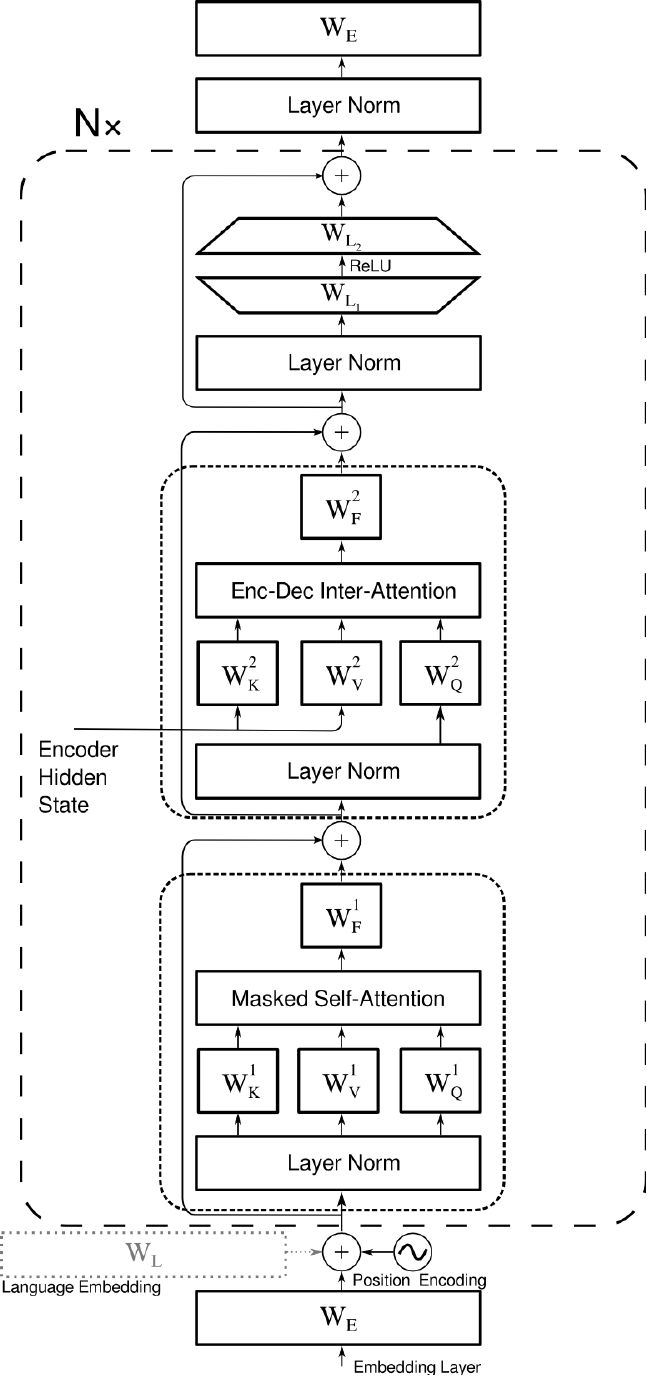}  
\caption{The decoder component of the transformer model \cite{vaswani-transformer:2017}. All parameters may be shared across all target tasks, or a unique set of decoder parameters can be created for each task (outer dashed line). Alternatively, we can create unique attention parameters for each task, while sharing the final feed-forward layers (inner dotted lines). The possibility of including an embedding for the target task is visualized at the bottom of the diagram. Illustration modeled after \citet{sachan-neubig-2018-parameter}.}
\label{fig:transformer_decoder}
\end{figure}

Our work builds upon recent research on many-to-one, one-to-many, and many-to-many translation models. We are interested in evaluating many-to-many models under realistic conditions, including:

\begin{enumerate}
    \item A highly imbalanced amount of training data available for different language pairs.
    \item A very diverse set of source and target languages.
    \item Training and evaluation data from many domains.
\end{enumerate}

\noindent We focus on multi-layer transformer models \cite{vaswani-transformer:2017}, which achieve state-of-the-art performance on large-scale MT and NLP tasks \cite{devlin2018bert,bojar-EtAl:2018:WMT1}. The decoder component of the transformer is visualized in figure \ref{fig:transformer_decoder}. We study four ways of building multi-lingual translation models. Importantly, all of the models we study can do zero-shot translation: translating between language pairs for which no parallel data was seen at training time. The models use training data from 11 distinct languages\footnote{\textsc{cs}, \textsc{de}, \textsc{en}, \textsc{fi}, \textsc{fr}, \textsc{gu}, \textsc{kk}, \textsc{lt}, \textsc{ru}, \textsc{tr} and \textsc{zh}}, with supervised data available from the WMT19 news-translation task for 22 of the 110 unique translation directions\footnote{Note we do not consider auto-encoding, thus the number of translation directions is $11^{2}-11=110$.}. This leaves 88 translation directions for which no parallel data is available. We try to evaluate zero-shot translation performance on all of these additional directions, using both gold parallel data, and evaluations based on pivoting or multi-hop translation. 

\paragraph{Target Language Specification}

Although the embedding and encoder parameters of a multi-lingual system may be shared across all languages without any special modification to the model, \textit{decoding} from a multi-lingual model requires a means of specifying the desired output language. Previous work has accomplished this in different ways, including: 

\begin{itemize}
    \item pre-pending a special target-language token to the input \cite{gnmt2016}
    \item using an additional embedding vector for the target language \cite{lample2019cross}
    \item using unique decoders for each target language \cite{luong2016,firat-etal-2016-multi}
    \item partially sharing some of the decoder parameters while keeping others unique to each target language \cite{sachan-neubig-2018-parameter,blackwood-etal-2018-multilingual}.
\end{itemize}

However, to the best of our knowledge, no side-by-side comparison of these approaches has been conducted. We therefore train models which are identical except for the way that decoding into different target languages is handled, and conduct a large-scale evaluation. We use only the language pairs and official parallel data released by the WMT task organisers, meaning that all of our systems correspond to the constrained setting of the WMT shared task, and our experimental settings should thus be straightforward to replicate. 

\section{Multi-Task Translation Models}

This section discusses the key components of the transformer-based NMT model, focusing on the various ways to enable translation into many target languages. We use the terms source/target \textit{\textbf{task}} and \textit{\textbf{language}} interchangeably, to emphasize our view that multi-lingual NMT is one instantiation of the more general case of multi-task sequence to sequence learning. 

\subsection{Shared Encoders and Embeddings}

In this work, we are only interested in ways of providing target task information to the model -- information about the source task is never given explicitly, and encoder parameters are always fully shared across all tasks. The segmentation model and embedding parameters are also shared between all source and target tasks (see below for more details).

\subsection{Multi-lingual Decoder Configurations}

Figure \ref{fig:transformer_decoder} visualizes the decoder component of the transformer model, with dashed and dotted lines indicating the parameter sets that we can replicate or share across target tasks.

\subsubsection{Target Task Tokens (\textsc{Prepend})}

\citet{gnmt2016} showed that, as long as a mechanism exists for specifying the target task, it is possible to share the decoder module's parameters across all tasks. In the case where all parameters are shared, the decoder model must learn to operate in a number of distinct modes which are triggered by some variation in the input. A simple way to achive this variation is by pre-pending a special "task-token" to each input. We refer to this method as \textbf{\textsc{Prepend}}.

\subsubsection{Task Embeddings (\textsc{Emb})}

An alternative to the use of a special task token is to treat the target task as an additional input feature, and to train a unique embedding for each target task \cite{lample2019cross}, which is combined with the source input. This technique has the advantage of explicitly decoupling target task information from source task input, introducing a relatively small number of additional parameters. This approach can also be seen as adding an additional token-level \textit{feature} which is the same for all tokens in a sequence \cite{sennrich-haddow:2016:WMT}. We refer to this setting as \textbf{\textsc{Emb}}.

\subsubsection{Task-specific Decoders (\textsc{Dec})}

In general, any subset of decoder parameters may be replicated for each target language, resulting in parameter sets which are specific to each target task. At one extreme, the entire decoder module may be replicated for each target language, a setting which we label \textbf{\textsc{Dec}} \cite{dong-etal-2015-multi}.

\subsubsection{Task-specific Attention (\textsc{Attn})}

An approach somewhere in-between \textsc{Emb} and \textsc{Dec} is to partially share some of the decoder parameters, while keeping others unique to each task. Recent work proposed creating unique attention modules for every target task, while sharing the other decoder parameters \cite{sachan-neubig-2018-parameter,blackwood-etal-2018-multilingual}. The implementation of their approaches differ significantly -- we propose to create completely unique attention parameters for each task. This means that for each of our 11 languages, we have unique context- and self-attention parameters in each layer of the transformer decoder. We refer to this setting as \textbf{\textsc{Attn}}.


\section{Experiments}

All experiments are conducted using the transformer-base configuration of \citet{vaswani-transformer:2017} with the relevant modifications for each system discussed in the previous section. We use a shared sentencepiece\footnote{\url{https://github.com/google/sentencepiece}} segmentation model with 32000 pieces. We use all available parallel data from the WMT19 news-translation task for training, with the exception of \texttt{commoncrawl}, which we found to be very noisy after manually checking a sample of the data, and \texttt{paracrawl}, which we use only for \textsc{en-fi} and \textsc{en-lt}\footnote{Turkish (\textsc{tr}) is included from the 2018 language pairs because the task-organizers suggest the possibility of using \textsc{tr} data to improve \textsc{kk} performance}. 

We train each model on two P100 GPUs with an individual batch size of up to 2048 tokens. Gradients are accumulated over 8 mini-batches and parameters are updated synchronously, meaning that our effective batch size is $2 * 2048 * 4 = 16384$ tokens per iteration. Because the task pair for each mini-batch is sampled according to our policy weights and (fixed) random seed, and each iteration consists of 8 unique mini-batches, a single parameter update can potentially contain information from up to 8 unique task pairs. We train each model for 100,000 iterations without early stopping, which takes about 40 hours per model. When evaluating we  always use the final model checkpoint (i.e. the model parameters saved after 100,000 iterations). We use our in-house research NMT system, which is heavily based upon OpenNMT-py \cite{opennmt}.

\insertDatasetInformationTable 

The sampling policy weights were specified manually by looking at the amount of available data for each pair, and estimating the difficulty of each translation direction. The result of the sampling policy is that lower resource language pairs are upsampled significantly. Table \ref{tab:dataset-information} summarizes the statistics for each language pair. Note that the data in each row represents a \textit{pair} of tasks, i.e. the total number of segments seen for \textsc{en-cs} is split evenly between \textsc{en}$\rightarrow$\textsc{cs}, and \textsc{cs}$\rightarrow$\textsc{en}. Because we train for only 100,000 iterations, we do not see all of the available training data for some high-resource language pairs.

With the exception of the \textsc{Prepend} system, the input to each model is identical. Each experimental setting is mutually exclusive, i.e. in the \textsc{Emb} setting we do not prepend task tokens, and in the \textsc{Attn} setting we do not use task embeddings.

\begin{figure}[!ht]  
\centering
\includegraphics[width=0.5\textwidth]{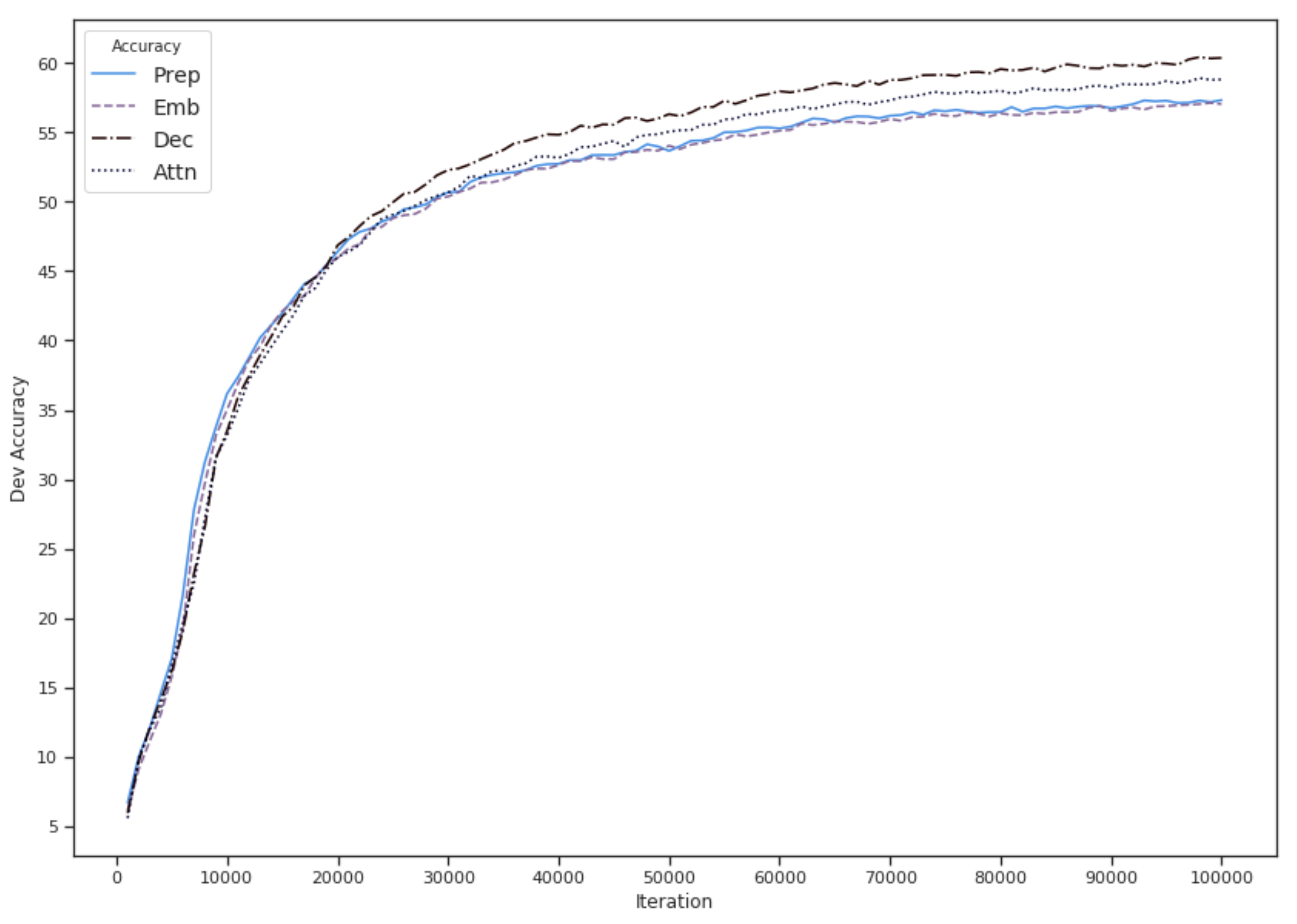}  
\caption{Word-level accuracy on WMT EN-DE 2014 dev set as training progresses. The model which has a \textsc{DE}-specific decoder achieves the highest accuracy on this dev set.} 
\label{fig:training-dev-progress}
\end{figure}

\begin{figure}[!ht]  
\centering
\includegraphics[width=0.5\textwidth]{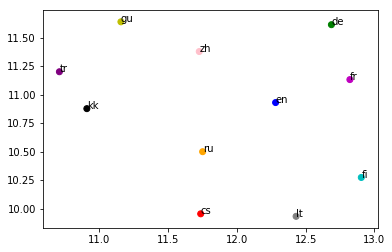}  
\caption{Language embeddings of the \textsc{Emb} system projected with UMAP \cite{mcinnes2018umap-software}.}
\label{fig:language-emb-visualization}
\end{figure}

Figure \ref{fig:training-dev-progress} plots the validation performance during training on one of our validation datasets. The language embeddings from the \textsc{Emb} system are visualized in figure \ref{fig:language-emb-visualization}.

\subsection{Results}

\insertEvaluationDatasetTable

We conduct four different evaluations of the performance of our models. First, we check performance on the 22 supervised pairs using dev and test sets from the WMT shared task. We then try to evaluate zero-shot translation performance in several ways. We use the TED talks multi-parallel dataset \cite{Ye2018WordEmbeddings} to create gold sets for all zero-shot pairs that occur in the TED talks corpus, and evaluate on those pairs. We also try two ways of evaluating zero-shot translation without gold data. In the first, we do round-trip translation $\textsc{SRC}\rightarrow \textsc{Pivot} \rightarrow \reallywidehat{\textsc{SRC}}$, and measure performance on the $(\reallywidehat{\textsc{SRC}}, \textsc{SRC})$ pair -- this method is labeled \textsc{Zero-Shot Pivot}. In the second, we use parallel evaluation datasets from the WMT shared tasks (consisting of $(\textsc{SRC}, \textsc{REF})$ pairs), and translate $\textsc{SRC} \rightarrow \textsc{Pivot} \rightarrow \reallywidehat{\textsc{TRG}}$, then measure performance on the resulting $ (\reallywidehat{\textsc{TRG}}, \textsc{REF}) $ pairs (see below for more details), where the pivot and target language pair is a zero-shot translation task -- this method is labeled \textsc{Zero-Shot Parallel Pivot}\footnote{For the \textsc{Zero-Shot Pivot} and \textsc{Zero-shot} parallel pivot evaluations we use the first 1000 segments of each dataset, because we need to translate twice for every possible pair.}

Table \ref{tab:evaluation-datasets} lists the WMT evaluation dataset that we use for each language pair. In the \textsc{Zero-Shot Pivot} setting, the reference side of the dataset is used as input.

Table \ref{tab:global-bleu} shows global results for all parallel tasks and all zero-shot tasks, by system. Global scores are obtained by concatenating the segmented outputs for each translation direction, and computing the BLEU score against the corresponding concatenated, segmented reference translations. The results in table \ref{tab:global-bleu} are thus \textit{tokenized} BLEU scores. 

\insertGlobalBleuTable

\subsection{Parallel Tasks}

Table \ref{tab:parallel-tasks} lists results for all supervised task pairs from WMT 2019. For each pair, we report BLEU scores on de-tokenized output, and compute scores using sacrebleu
\footnote{\texttt{BLEU+case.mixed+ lang.<src-lang>-<trg-lang>+ numrefs.1+smooth.exp+tok.<trg-lang>+ version.1.2.19}}. Therefore, we expect BLEU scores to be equivalent to those used in the WMT automatic evaluation. 

\insertParallelTaskTable

We note that across all but the lowest-resource tasks, the model with a unique decoder for each language outperforms all others. However, for \textsc{en$\rightarrow$gu} and \textsc{en$\rightarrow$kk}, the lowest-resource translation directions, the unique decoder model fails completely, probably because the unique parameters for \textsc{kk} and \textsc{gu} were not updated by a sufficient number of mini-batches (approximately 15,600 for \textsc{en$\rightarrow$gu} and 14,800 for \textsc{en$\rightarrow$kk}).

\subsection{Zero-shot Translation Tasks}

\insertShortZeroshotTable

In order to test our models in the zero-shot setting, we first create a multi-parallel dataset from the from the TED Talks multi-parallel corpus \cite{Ye2018WordEmbeddings}, which has recently been used for the training and evaluation of multi-lingual models. We filter the dev and test sets of this corpus to find segments which have translations for all of \textsc{en, fr, ru, tr, de, cs, lt, fi}, and are at least 20 characters long, resulting in 606 segments. Because this corpus is pre-processed, we first de-tokenize and de-escape punctuation using \texttt{sacremoses}\footnote{https://github.com/alvations/sacremoses}. We then evaluate zero-shot translation for all possible pairs which do not occur in our parallel training data, aggregate results are shown in the second row of table \ref{tab:global-bleu}. 

We then adapt an evaluation technique that has recently been used for unsupervised MT -- we translate from the source language into a pivot language, then back into the source language, and evaluate the score of the resulting source-language hypotheses against the original source \cite{lampleCDR18}. This technique allows us to evaluate for all possible translation directions in our multi-directional model. 

\insertFailedTasksTable

Aware of the risk that the model simply copies through the original source segment instead of translating, we assert that at least 95\%  of pivot translations' language code is correctly detected by  \texttt{langid}\footnote{https://github.com/saffsd/langid.py}, and pairs which do not meet this criteria for any system are removed from the evaluation for all systems (not just for the system that failed). For all models except \textsc{Emb} only \textsc{ru$\rightarrow$kk$\rightarrow$ru} \textsc{fi$\rightarrow$lt$\rightarrow$fi}, and \textsc{zh$\rightarrow$gu$\rightarrow$zh} failed this test, but for the \textsc{Emb} model 31 of the 110 translation directions failed (see tables \ref{tab:failed-pivot-tasks} and \ref{tab:pivot-translation-full-results}(in appendix)\footnote{We conduct round trip translation on all 110 directions, but we only use directions that are (1) not available in the parallel training data, and (2) pass the language identification test to compute the global zero-shot translation performance.}. This result indicates that models which use language embeddings may have a more "fuzzy" representation of the output task, and are much more prone to copying than other approaches to multi-lingual MT. 

Finally, we conduct the \textsc{Zero-Shot Parallel Pivot} evaluation using the same datasets in table \ref{tab:evaluation-datasets} by translating from \textsc{EN} (or \textsc{DE} in the case of \textsc{*-FR}) to each possible pivot language, and then from the pivot language into the target language. Compared to the \textsc{Zero-Shot Pivot} setting, this evaluation should help to protect against the risk of copying, because source and reference segments are not from the same language. Aggregate results for this setting are shown in the third row of table \ref{tab:global-bleu}, full results in table \ref{tab:parallel-pivot-translation-full-results} in appendix. 

\subsection{Discussion}

Our results show that a models with either (1) a completely unique decoders for each target language or (2) unique decoder attention parameters for each target language clearly outperform models with fully shared decoder parameters. 

As shown in table \ref{tab:global-bleu}, the \textsc{Zero-Shot Pivot} evaluation is the outlier in our results, with the \textsc{Emb} system outperforming the others. Even for the languages which passed the language identification filter used in this evaluation, we suspect that some copying is occurring for the \textsc{Emb} system, because of the mismatch in results between the \textsc{Zero-Shot Pivot} task and the \textsc{Supervised, Zero-Shot TED}, and \textsc{Zero-shot Parallel Pivot} tasks (see table \ref{tab:global-bleu}). Since the ranking of the models according to the \textsc{Zero-Shot Parallel Pivot} evaluation is well aligned with the \textsc{Zero-Shot TED} and \textsc{Supervised} evaluations which use gold parallel evaluation data, we believe that this method is effective for zero-shot evaluation of translation quality for language pairs where no gold data is available.

It is plausible that the language-independence of encoder output could be correlated with the amount of sharing in the decoder module. Because most non-English target tasks only have parallel training data in English, a unique decoder for those tasks only needs to learn to decode from English, not from every possible source task. However, our results show that the \textsc{Attn} model, which partially shares parameters across target languages only slightly outperforms the \textsc{Dec} model globally,
because of the improved performance of the \textsc{Attn} model on the lowest-resource tasks (Table \ref{tab:parallel-tasks}, Table  \ref{tab:pivot-translation-full-results} (in appendix)). Thus we conclude that multi-lingual encoders still learn to share information across languages, even when trained using decoders that are unique to each target task.

\section{Related Work}

\citet{dong-etal-2015-multi,firat-etal-2016-multi,ha-multilingual-2016,johnson-google-2016} and others have shown that multi-way NMT systems can be created with minimal modification to the approach used for single-language-pair systems. \citet{johnson-google-2016} showed that simply prepending a target-task token to source inputs is enough to enable zero-shot translation between language pairs for which no parallel training data is available. 

Our work is most similar to \citet{sachan-neubig-2018-parameter}, where several different strategies for sharing decoder parameters are investigated for one-to-many translation models. However, their evaluation setting is constrained to one-to-many models which translate from English into two target languages, whereas our setting is more ambitious, performing multi-way translation between 11 languages. \citet{blackwood-etal-2018-multilingual} showed that using separate attention parameters for each task can improve the performance of multi-task MT models -- this work was the inspiration for the \textsc{Attn} setting in our experiments. 

Several recent papers focus specifically upon improving the zero-shot performance of multi-lingual MT models \cite{chen-etal-2017-teacher,arivazhagan2019missing,gu2019improved,lu-etal-2018-neural,Maruan-consistency,sestorain2019zeroshot}. 

Concurrently with this work, \cite{aharoni-etal-2019-massively} evaluated a multiway MT system on a large number of language pairs using the TED talks corpus. However, they focus upon \textsc{EN-*} and \textsc{*-EN}, and do not test different model variants. 

\section{Conclusions and Future Work}

We have presented results which are consistent with recent smaller-scale evaluations of multi-lingual MT systems, showing that assigning unique attention parameters to each target language in a multi-lingual NMT system is optimal when evaluating such a system globally. However, when evaluated on the individual task level, models which have unique decoder parameters for every target task tend to outperform other configurations, except when the amount of available training data is extremely small. We have also introduced two  methods of evaluating zero-shot translation performance when parallel data is not available, and we conducted a large-scale evaluation of translation performance across all possible translation directions in the constrained setting of the WMT19 news-translation task. 

In future work, we hope to continue studying how multi-lingual translation systems scale to realistic volumes of training data and large numbers of source and target tasks.



\bibliography{acl2019}
\bibliographystyle{acl_natbib}

\appendix

\insertMTMatrixZeroshotTable

\insertParallelPivotMTMatrixTable

\end{document}